\documentclass[nofootinbib]{elsarticle}
\usepackage[margin=1in]{geometry}
\usepackage{amsmath}
\usepackage{graphicx}% Include figure files
\usepackage{subcaption}
\usepackage{booktabs} % For professional quality tables (\toprule, \midrule, \bottomrule)
\usepackage{multirow} % For cells that span multiple rows
\usepackage{array}    % For more advanced column specifications (though not strictly used in complex ways here)
\usepackage{algorithm}
\usepackage{algpseudocode}%%
\usepackage{hyperref} % make references clickable links
\usepackage{hvfloat}
\usepackage{xfrac}
\usepackage{url}
\usepackage{hyperref}
\usepackage{natbib,hyperref}
\usepackage{doi} 
\AtBeginDocument{%
  }

\begin{document}

\title{a-TMFG: Scalable Triangulated Maximally Filtered Graphs via Approximate Nearest Neighbors}

\author[uct-sta]{Lionel Yelibi}
% [orcid=0000-0000-XXXX-XXXX]
\ead{ylblio001@myuct.ac.za}
% ACM style
%\affiliation{%
%  \institution{University of Cape Town}
%  \city{Rondebosch, Cape Town}
%  \state{Western Cape}
%  \country{South Africa}
%}
\address[uct-sta]{Department of Statistical Sciences, University of Cape Town,\\
Rondebosch, 7701, Cape Town, Western Cape,
South Africa}

% \affiliation{%

% \affiliation{%

% \renewcommand{\shortauthors}{Lionel Yelibi}

\begin{abstract}
The traditional Triangular Maximally Filtered Graph (TMFG) construction requires pre-computation and storage of a dense correlation matrix; this limits its applicability to small and medium-sized datasets. Here we identify key memory and runtime complexity challenges when using TMFG at scale. We then present the Approximate Triangular Maximally Filtered Graph (a-TMFG) algorithm. This is a novel approach to scaling the construction of artificial graphs from data inspired by TMFG. The method employs k-Nearest Neighbors Graphs (kNNG) for initial construction, and implements a memory management strategy to search and estimate missing correlations on-the-fly. This provides representations to control combinatorial explosion. The algorithm is tested for robustness to the parameters and noise, and is evaluated on datasets with millions of observations. This new method provides a parsimonious way to construct graphs for use-cases where graphs are used as input to supervised and unsupervised learning but where no natural graph exists. 
\end{abstract}

\maketitle
\tableofcontents

\section{Introduction}

Modern statistical learning often boils down to finding a more informative representation for data to allow one to  compute a meaningful distance between observations. These distances, when mapped to a graph can then be used in many machine learning tasks. In both unsupervised learning: such as graph clustering, manifold learning, and generative modeling, but also graph supervised learning tasks: such as node prediction, link prediction, and graph classification. The `graph revolution' is seen across industries such as financial services with payment networks, utilities with electricity networks, transportation with roads and airport networks and more\cite{paul_systematic_2024,he_overview_2021,liao_review_2022,wang_review_2022,zhou_graph_2020}. % cite something about gnn usecases.

The majority of data in most corporate environments is tabular with no obvious associated graphs. This has limited the usefulness of graph predictive models to problems for whom a natural graph exists. This motivates the search for methods which would allow the construction of graphs learned from data with desirable properties for machine learning tasks. In healthcare previous work \cite{muller_survey_2023} regularization techniques were introduced to constrain graph construction to deal with density. This allows one to introduce parsimony via local and global optimization methods such as Locally Linear Embedding (LLE) \cite{ghojogh_locally_2020,roweis_nonlinear_2000,saul_introduction_nodate}, or the famous Graphical LASSO \cite{friedman_sparse_2008}, which produce sparse correlation matrices. Variations of these graph construction techniques involve first finding a reduced representation of the data on a lower manifold. This is either done by embracing a linear decomposition of the data which can either be local with observations features decomposed as linear combination of a reduced set of other observations, or global where the observation features are linear combination of a reduced set of common factors influencing all observations in the dataset. Although local methods sound attractive they suffer from the curse of dimensionality as they require solving $N$ equations with $N$ unknown coefficients augmented with a regularization term to produce a sparse representation which has the effect of severely limiting their scalability.

One approach the imposing a graph onto data was to consider Minimum Spanning Tree (MST) representations \cite{f_bazlamacci_minimum-weight_2001}. The MST is a special kind of graph which takes the form of a tree connecting all nodes of a graph, minimizes the total path length of the tree, and produces a connected graph. The MST has however been shown susceptible to noise \cite{israeli_noise_2024} and is not dense enough to capture richer topological features which may be relevant. 

The Triangular Maximally Filtered Graph (TMFG) provides a viable and more general approach \cite{previde_massara_building_2020,massara_learning_2021,massara_network_2017,tumminello_tool_2005}; this is an approximate solution to the more general problem of the Planar Maximum Weight subgraph \cite{calinescu_maximum_2018}. It always contains the MST which makes it by construction a connected graph and whose edgeset has a fixed size of $3N - 6$ which caps its memory complexity to $N$. The sparsity induced by the TMFG has made it a desirable candidate to model collective behavior in stock market securities, and features ranking in unsupervised feature selection \cite{briola_topological_2023}. 

However, the technique requires the pre-computation of $N^2$ correlation coefficients which has restricted its use to small scale problems. Parallelized versions of the TMFG exist which allow its accelerated construction by adding multiple nodes at the same time which produces significant speed \cite{yu_parallel_2023,yu_parallel_2024,raphael_faster_2024}, or by using priority queues \cite{noauthor_ultra-fast_2024}. These techniques still require a dense correlation matrix as input, and thus these improvements remain limited in practice because of the scaling constraints. 

Here we draw inspiration from the foundational principles of the TMFG to create a topologically similar graph, but with strictly controlled memory and runtime complexities. We introduce this novel method as the Approximate Triangular Maximally Filtered Graph (a-TMFG). By utilizing a $k$-NN graph to guide the initial traversal, estimating missing correlations on-the-fly, and bounding the active pool of candidate cliques, a-TMFG efficiently produces maximal planar graphs for datasets scaling into the hundreds of thousands of observations. 

The remainder of this paper is structured as follows: Section \ref{sec:scalability} introduces the mechanics of the exact TMFG and discusses the architectural limitations that prevent its scalability. Our novel a-TMFG algorithm and its optimizations are detailed in Section \ref{sec:algorithm}. In Section \ref{sec:evaluation}, we evaluate the algorithm's ability to recover ground-truth edges using synthetic Gaussian Markov Random Fields (GMRF) and analyze the impact of key hyper-parameters on performance and runtime. Finally, Section \ref{sec:discussion} highlights the method's empirical advantages, acknowledges current limitations, and outlines directions for future research.

\section{Scalability and Motivation} \label{sec:scalability}

A set of three interconnected nodes $v_j$, $ v_{j+1}$, and $v_{j+2}$ form a face: $f_j=(v_j, v_{j+1}, v_{j+2})$. This is the $j$-th face and is also known as a 3-clique, or triangle face. These are the ``faces'' of the graph. The TMFG is constructed by connecting these 3-cliques. Concretely, the graph grows by connecting a new node $v_k$ (the $k$-th node) to an existing face while strictly maintaining planarity. Geometrically, the addition of a new node $v_k$ to a face $f_j = (v_j, v_{j+1}, v_{j+2})$ forms a tetrahedron. This process partitions the space into three new interior faces: $f_{j+1} = (v_k, v_{j+1}, v_{j+2})$, $f_{j+2} = (v_j, v_k, v_{j+2})$ and $f_{j+3} = (v_j, v_{j+1}, v_k)$. The original face $f_j$ is ``covered" by the new node $v_k$ and removed from the active set of potential attachment points. 

This process repeats until all $N$ nodes are integrated, resulting in exactly $3N-6$ edges. 
Because each iteration adds three faces $\{ f_{j+1}, f_{j+2},f_{j+3} \}$ and deletes one $\{f_{j}\}$, the total face universe $\mathcal{F}$ grows by two per step, eventually reaching a set of $2N-4$ active faces. $\mathcal{F}$ is the set of candidate faces which can add new nodes at any given time. Each face $f_j$ is scored against the set of nodes yet to be added to the graph which is $\mathcal{O}(N)$. Thus the memory complexity of the face universe $\mathcal{F}$ is $\mathcal{O}(N^2)$ and its computing complexity is $\mathcal{O}(N^2)$. In both cases the cost of storing and computing the needed quantities to construct the TMFG renders it unscalable when $N$ grows beyond the tens of thousands.

\begin{figure}[h]
    \centering
    \includegraphics[width=.75\linewidth]{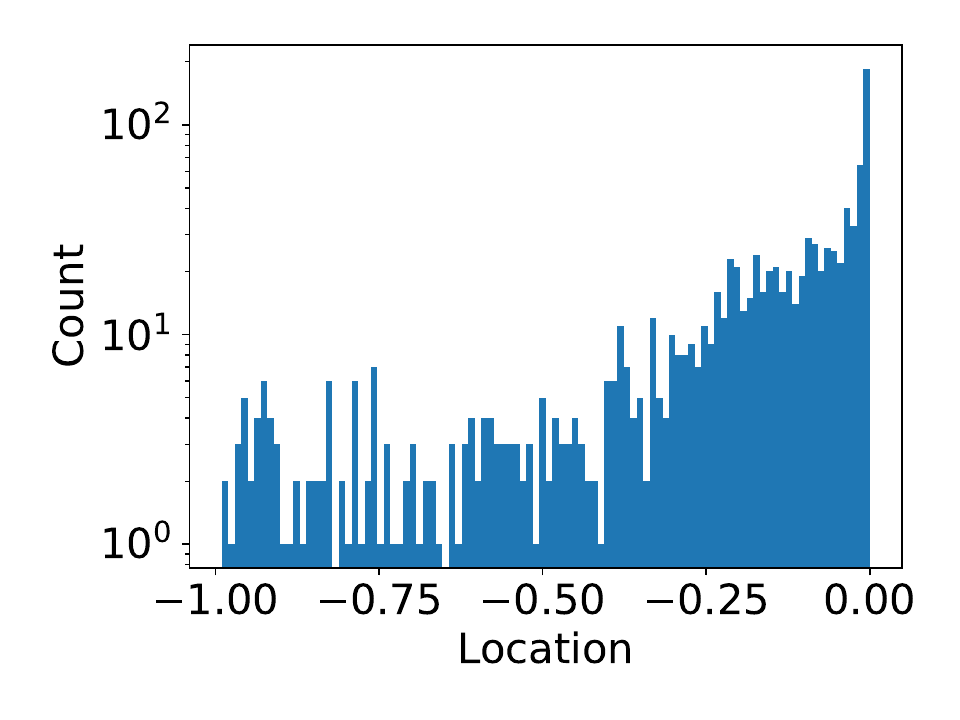}
    \caption{The face counts against the location $L_j$ of a given face $f_j$ in the face universe $\mathcal{F}$ used in the TFMG algorithm. This shows that most of the selected faces are found at the end of the sequence of faces (on the right near location $L_j=0$) rather than near the beginning (near location $L_j=-1$). This implies that scaling can be enhanced by exploiting the exploration frontier on the right of the plot, and forgetting unnecessary faces (See Section \ref{sec:scalability}). This is what the ``approximate" refinement of the TMFG method (a-TFMG) exploits.}

    % \caption{Distribution of the relative selection location $L_j$ for faces $f_j$ within the active universe $\mathcal{F}$. Selections overwhelmingly occur near the exploration frontier ($L_j \approx 0$), empirically motivating the bounded universe optimization introduced in Section \ref{sec:scalability}.}
    \label{fig:facestats}
\end{figure}

To better understand this scaling problem we need to look at the location $L_j$ of the selected face $f_j$ in universe $\mathcal{F}$: $L_j = \tfrac{1}{|\mathcal{F}|}{(j - |\mathcal{F}|)}$.
Here $j$ is the index of the selected face $f_j$. The index is chosen so that if it is located at the end of $\mathcal{F}$, then $L_j = 0$, and if at the beginning of $\mathcal{F}$, $L_j = -1$. In Figure \eqref{fig:facestats} we show the distribution of $L_j$ for a synthetic example. Here most selected faces are located near the end of $\mathcal{F}$ ({\it i.e.} $L_j = 0$) suggesting the existence of something akin to an exploration frontier. Because most faces selected are located near the frontier, this also suggests one can ``forget'' the earlier faces in $\mathcal{F}$ to at least halve its size, without severely impacting the quality of the final graph. This is the key insight we exploit here. 

\section{Approximate Triangular Maximally Filtered Graphs (a-TMFG)} \label{sec:algorithm}

To address traditional computational bottlenecks, we propose an architecture that transitions the algorithm from an unmanageable ${\cal O}(N^2)$ complexity toward an efficient, highly scalable profile. As outlined in Algorithm \ref{alg:high_level}, this is achieved through three primary mechanisms:

\begin{enumerate}
    \item Approximate Nearest Neighbor Indexing: By substituting the exhaustive correlation matrix with a Hierarchical Navigable Small World (HNSW)\cite{malkov_efficient_2020} index ($\mathcal{I}$) and a static sparse graph ($G$), we heavily reduce the search space complexity.
    \item Bounded Universe and Priority Queue: We limit the maximum number of active faces ($\mathcal{F}$) stored in memory. Candidates are evaluated and pushed to a priority queue ($\mathcal{Q}$). By utilizing lazy deletion during the extraction phase, we skip outdated edges in $\mathcal{O}(1)$ time, reducing the scoring complexity to approximately $\mathcal{O}(U \times N)$, where $U \ll N$.
    \item Centroid Caching and Rescue: To prevent redundant mathematical operations, the sum-vector (centroid) of each face is computed once upon creation and cached. If the local search stalls, these cached vectors are passed directly to the HNSW index, which seamlessly skips integrated nodes to find the next optimal frontier connections.
\end{enumerate}

The proposed construction begins in Line 1 of Algorithm \ref{alg:high_level} by initializing the pre-computed kNN graph $G$ alongside its parent HNSW index $\mathcal{I}$. 

% --- Algorithm Block ---

\begin{algorithm}[h]
\caption{Approximate TMFG (a-TMFG)}\label{alg:high_level}
\begin{algorithmic}[1]
\Require Dataset $X$, Parameters $k$, clique size $c$
\State Build approximate nearest neighbor index $\mathcal{I}$ and sparse graph $G$ from $X$
\State Initialize Seed Clique $f_0$ from $G$
\State Initialize Priority Queue $\mathcal{Q}$ with local candidates around $f_0$
\State Initialize active faces $\mathcal{F}$
\While{Graph edges $<$ Target Edges}
    \State $(f_{\text{best}}, v_{\text{best}}) \gets \text{Pop}(\mathcal{Q})$
    \If{Local search exhausted ($v_{\text{best}}$ is null)}
        \State \Comment{\textbf{Global Rescue Phase}}
        \State Candidates $V' \gets \text{Query } \mathcal{I} \text{ using active faces } \mathcal{F}$
        \State Score $V'$ against $\mathcal{F}$ and push best pairs to $\mathcal{Q}$
    \Else
        \State \Comment{\textbf{Local Expansion Phase}}
        \State Connect $v_{\text{best}}$ to $f_{\text{best}}$ in Graph
        \State Remove $v_{\text{best}}$ from $\mathcal{I}$ (Mark as dead)
        \State Replace $f_{\text{best}}$ with new subdivided faces
        \State Find local neighbors for new faces using $G$, score, and push to $\mathcal{Q}$
    \EndIf
\EndWhile
\State \Return Edgelist
\end{algorithmic}
\end{algorithm}

While traditional TMFG implementations perform a global search over the full correlation matrix to identify the initial tetrahedron, we mitigate this overhead by querying the sparse graph $G$. In Lines 2 through 4, we identify a seed node whose nearest neighbors maximize the local sum of edge weights to form the seed clique $f_0$. We populate the edge list with the internal connections of this clique and generate the initial active faces of $\mathcal{F}$. For each new face, its centroid vector is calculated, cached, and used to score the union of its neighbors. The highest-scoring valid candidates are then pushed into the priority queue $\mathcal{Q}$.

The main construction iterates until the graph satisfies the maximal planar condition of $3N - 6$ edges (Line 5). At the start of each iteration (Line 6), the algorithm pops the highest-scoring candidate edge $(f^*, v^*)$ from $\mathcal{Q}$. Through a lazy-deletion mechanism, if the face has been previously pruned or the target node has already been integrated into the graph, the edge is discarded, and the algorithm retrieves the next valid candidate.

Assuming a valid node is identified, the algorithm proceeds to the \textit{Local Expansion Phase} (Lines 13--17). The optimal node $v^*$ is connected to all vertices of $f^*$. Crucially, the node is flagged in a global integration mask and marked as deleted in the HNSW index $\mathcal{I}$ (Line 15)and $f^*$ is then removed from $\mathcal{F}$. If the active face count exceeds a defined universe limit $U$, the oldest active faces are pruned. Finally, $f^*$ is subdivided into three new faces (Line 16). The algorithm retrieves their local neighbors from $G$, scores them using the newly cached centroid vectors, and pushes the new optimal pairs into $\mathcal{Q}$ (Line 17).

\textit{Global Rescue Phase:} A significant challenge with localized expansion is that the sparse graph $G$ may contain disconnected components, or local neighborhoods may naturally exhaust before the manifold is complete. In such scenarios, $\mathcal{Q}$ is depleted, and the pop operation returns a null value (Line 7). The cached centroid vectors of all currently active faces in $\mathcal{F}$ are queried against the HNSW index $\mathcal{I}$ in a single batch (Line 9). Because integrated nodes were continuously marked as deleted during the expansion phase (Line 15), the HNSW index automatically bypasses the interior of the graph, returning only the nearest unadded frontier nodes $V'$. These candidates are scored against the active faces, and the optimal bridging edges are injected back into $\mathcal{Q}$ (Line 10). This dynamically jump-starts the exploration, allowing the triangulation to bridge disconnected components and guaranteeing a connected graph.

\section{Evaluation} \label{sec:evaluation}

To evaluate graph construction algorithms with topological constraints such as the TMFG we need to generate synthetic data for which we know a ground truth solution. So we have two problems: 
\begin{enumerate}
    \item Generate a ground truth graph $G_t$
    \item Generate data $X = f(G_t)$
\end{enumerate}
Mainstream graph generator functions provided by network science python libraries such as networkx and igraph are not able to generate synthetic TMFGs, especially at scale. They also only generate graphs alone without node features. The next candidate is to generate data from a factor model (i.e. Gaussian Mixtures) with a known sparse correlation matrix\footnote{This is achieved with a model as follows:
$$ X_i = g_s \eta_s + \sqrt{ 1 - g_s^2 } \epsilon_i $$
where $X_i$ are the data features, $g_i$ the factor loading which controls correlation of object $i$ to the factor $\eta_s$, and the object's idiosyncratic noise $\epsilon_i$ independent from the factor. Both $\eta$ and $\epsilon$ are typically chosen to be standard normal.}. There are $K$ factors representing individual non-overlapping clusters such that each object $i$ is only influenced by a single $\eta$ factor. 

\begin{figure}[H]
    \centering
    
    \begin{subfigure}{0.48\linewidth}
        \centering
        \includegraphics[width=\linewidth]{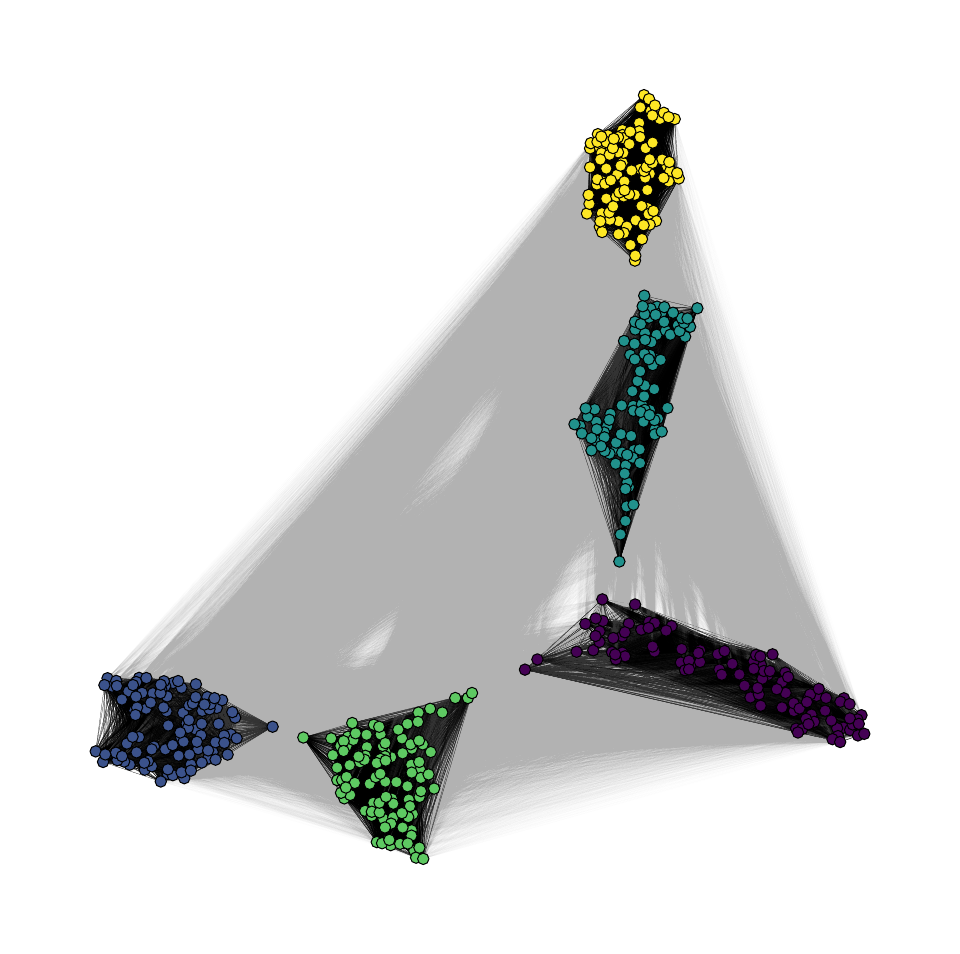}
        \caption{Block Diagonal Graph}
        \label{fig:factor_graph}
    \end{subfigure}
    \hfill
    \begin{subfigure}{0.48\linewidth}
        \centering
        \includegraphics[width=\linewidth]{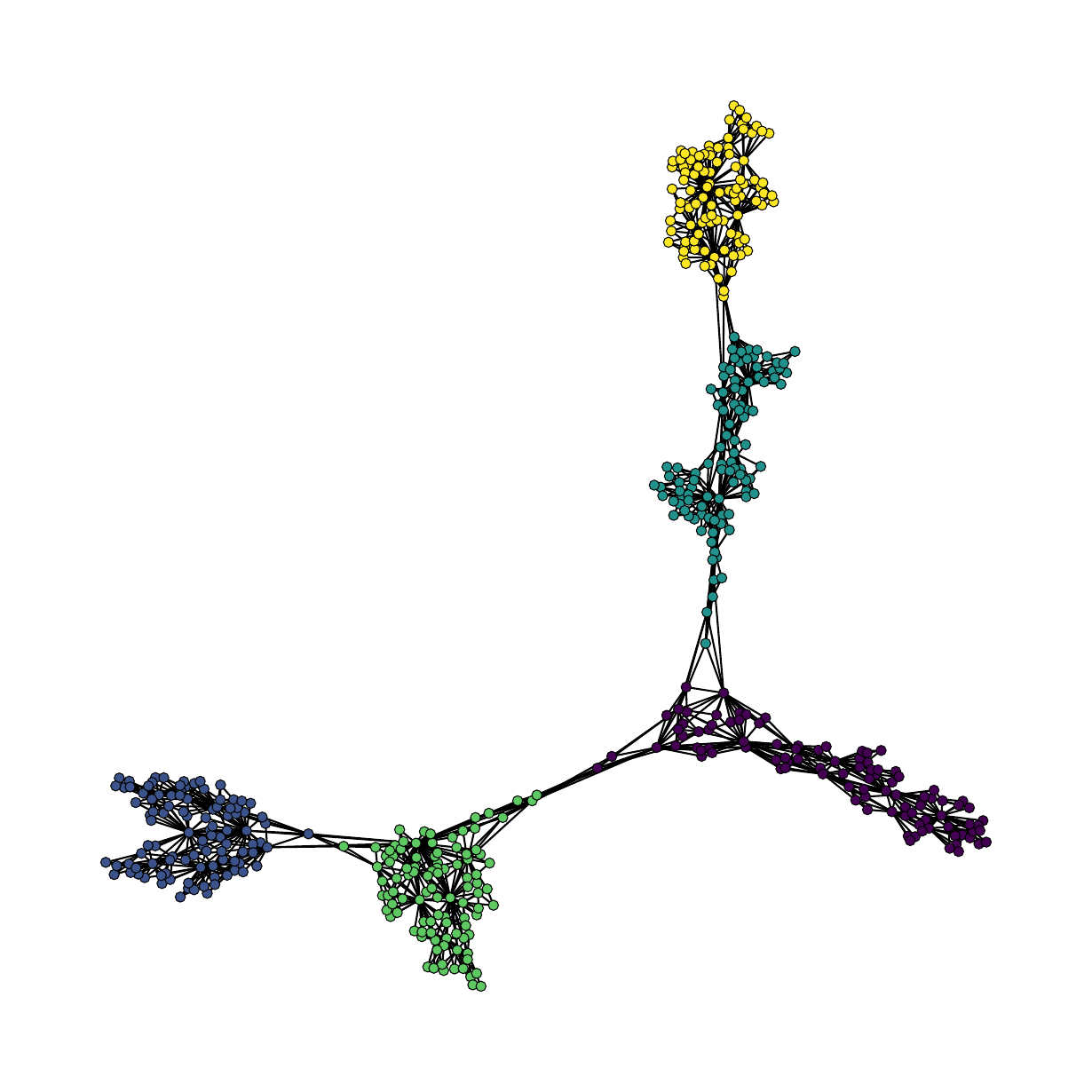}
        \caption{TMFG}
        \label{fig:factor_tmfg}
    \end{subfigure}
    \caption{Network topologies derived from a 1-factor Gaussian mixture model. (\subref{fig:factor_graph}) block-diagonal correlation matrix. (\subref{fig:factor_tmfg}) Exact TMFG constructed from the same dataset. The TMFG induces a sparse, hierarchical structure, highlighting the need for generative models capable of capturing short-range interactions (see Section \ref{sec:evaluation}).}
    \label{fig:motiv_graphs}
\end{figure}

Here we consider that we have a graph $G(V, E)$ with a set of nodes $V$, and edges $E$ such that the correlation between nodes of the same cluster $s$ is $g_s^2$ and $0$ otherwise. The true graph adjacency matrix is block diagonal but as a consequence of estimation noise a large number of non zero entries outside of clusters can be observed. This is observable in Figure \eqref{fig:motiv_graphs}: In Fig. \eqref{fig:factor_graph} we show the graph of a dataset generated by our 1-factor model where we highlight the edges in black when they connected nodes from the same clusters, and in gray other wise. In Fig. \eqref{fig:factor_tmfg} we show the TMFG estimated from the same dataset which produces a sparse graph. 

\begin{figure}[H]
    \centering
    
    \begin{subfigure}{0.45\linewidth}
        \centering
        \includegraphics[width=\linewidth]{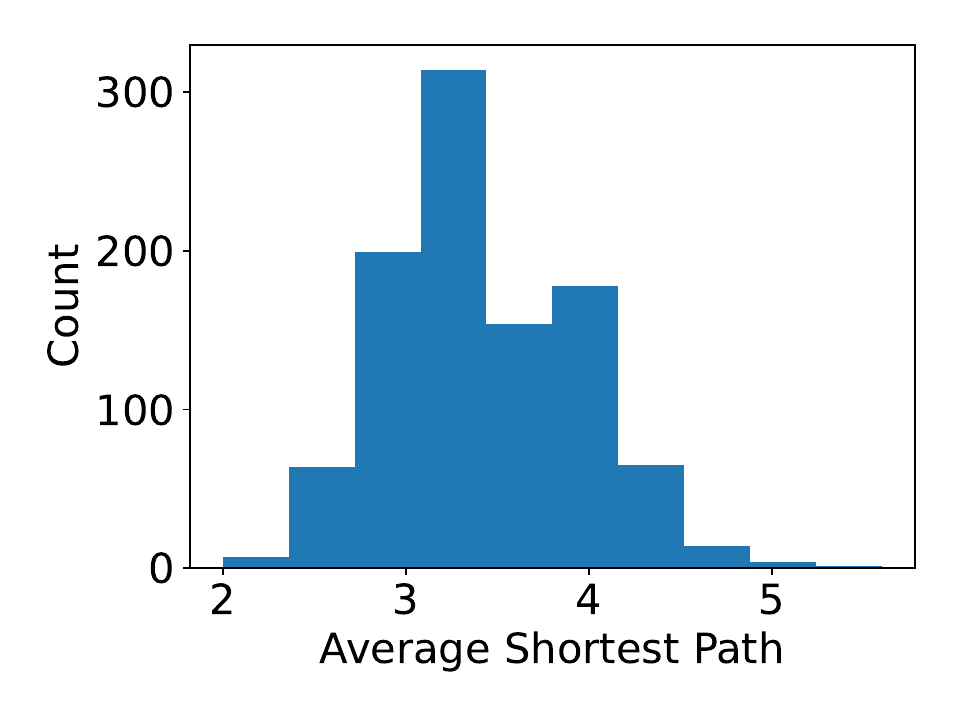}
        \caption{Shortest Path Distribution}
        \label{fig:dgp_short}
    \end{subfigure}
    ~
    \begin{subfigure}{0.45\linewidth}
        \centering
        \includegraphics[width=\linewidth]{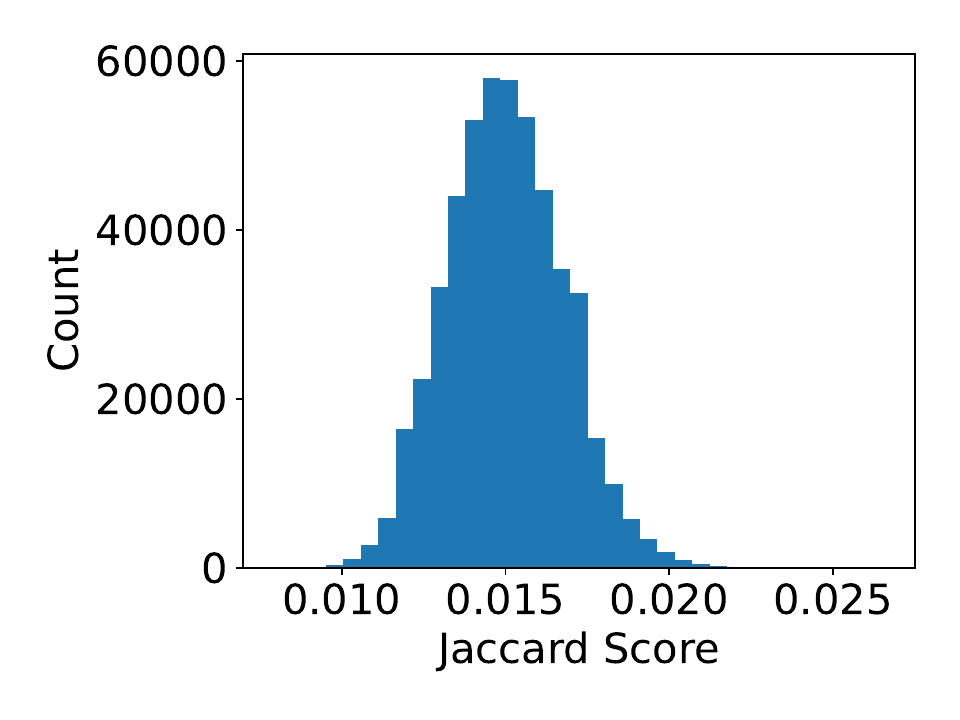}
        \caption{Jaccard Score Distribution}
        \label{fig:dgp_jac}
    \end{subfigure}
    \caption{Topological consistency of 1,000 exact TMFGs estimated from identically parameterized 1-factor models ($N=1000, K=5$). (\subref{fig:dgp_short})  Average shortest path between intra-cluster nodes. (\subref{fig:dgp_jac}) Distribution of pairwise Jaccard scores between the resulting edge lists. The near-zero Jaccard overlap illustrates the instability of TMFGs on purely block-diagonal matrices, motivating the use of GMRFs in Section \ref{sec:evaluation}.}
    \label{fig:motiv_stats}
\end{figure}

For multiple realization of the data generative process, the intra-cluster edges are theoretically identical but estimation noise leads greedy algorithms such as the TMFG to produce different outputs every single time. To illustrate this we generate one thousand correlation matrices from the same 1-factor model with parameters $N=1000$ samples, $K=5$ clusters, $g_s = 0.5$ and we compute their respective TMFG. In Fig. \eqref{fig:dgp_short} we show the average shortest path between two nodes of the same class with a mean 3 steps for clusters of size $\frac{N}{K} = 200$. The constructed TMFG graphs are able to capture a desirable property of the original correlation matrix: its block diagonal clustered structure. Next we want to understand whether the similarities between the edgelists of the TMFGs computed. We calculate the pairwise Jaccard Scores in Fig. \eqref{fig:dgp_jac} and we observe that the distribution is centered in 0 meaning there is virtually no overlap between the individual TMFG computed from factor model. Our argument has so far been that block diagonal correlation matrices are inappropriate to generate simulated data for the evaluation of TMFG algorithms. We have shown that despite generating data from the same identical factor model we are unable to recover identical TMFG solutions suggesting that the interaction in the original model are not short range interactions. In this specific case the dependencies are latent with respect to the factors which explain the observed behavior\footnote{This experiment can also serve as a rejection test for the TMFG on dynamical data: Consider a dataset measured in time. If a TMFG is computed at time $t_i$ and doesn't have a high overlap with another TMFG computed at time $t_{i+\tau}$ one could conclude that either the system is non-stationary or if stationary, the TMFG is inappropriate to model its dependencies.}.

\subsection{Gaussian Markov Random Field} \label{sec:gmrf}

In the Gaussian Markov Random Field \cite{rue_gaussian_2005} framework, we two sets of variables $X_i$ and $Y_i$, $X_i$ variables are idiosyncratic noise to the nodes on the graph $G(V,E)$, and $Y_i$ are functions of the graph's nodes interacting with their neighboring nodes such that 
$$Y_i = X_i + \sum_{j \in n_i} X_j$$
With this model $Y_i$ variables at every realization of the system are independent in time (IID) and the resulting system is multivariate normal distributed when $X_i$ are gaussian variables. The Gaussian Markov Random Field allows us to encode any graphical dependency structure which is a generalization of the classical 1-factor model we have previously used in \cite{yelibi_agglomerative_2021,yelibi_fast_2020} and in this paper. While the 1-factor model translates to a block diagonal adjacency matrix, we can encode flexible connectivity patterns with GMRF models. In our case we are interested in short range random processes over the graph which are suitable when the graph is a TMFG graph.

The Covariance matrix of such model is $$ C =  \left( I + A \right) \left( I + A \right)^{\top} = I + A + A^{\top} + AA^{\top} $$ where $A^2$ non-zero entries include interaction with neighbors of neighbors (2-hop connections). Instead we can define $C = LL^T$ with $L = \sqrt{ I + \alpha A}$ which can be approximated to $L = I + \frac{\alpha}{2} A $ such that 

\begin{equation} \label{eq:gmrf_cov}
    C =  \left(I + \frac{\alpha}{2} A\right)\left(I + \frac{\alpha}{2} A\right)^{\top} = I + \alpha A + \frac{\alpha^2}{4}A^2
\end{equation}

When $\alpha$ is low (i.e. $\alpha \leq 2 $) we suppress 2-hop connections due to $\frac{\alpha^2}{4} \leq 1 $. We then investigate the impact of $\alpha$ on the ability for a-TMFG to recover the right dependencies.

\begin{figure*}[h]
    \centering
    \includegraphics[width=.7\textwidth]{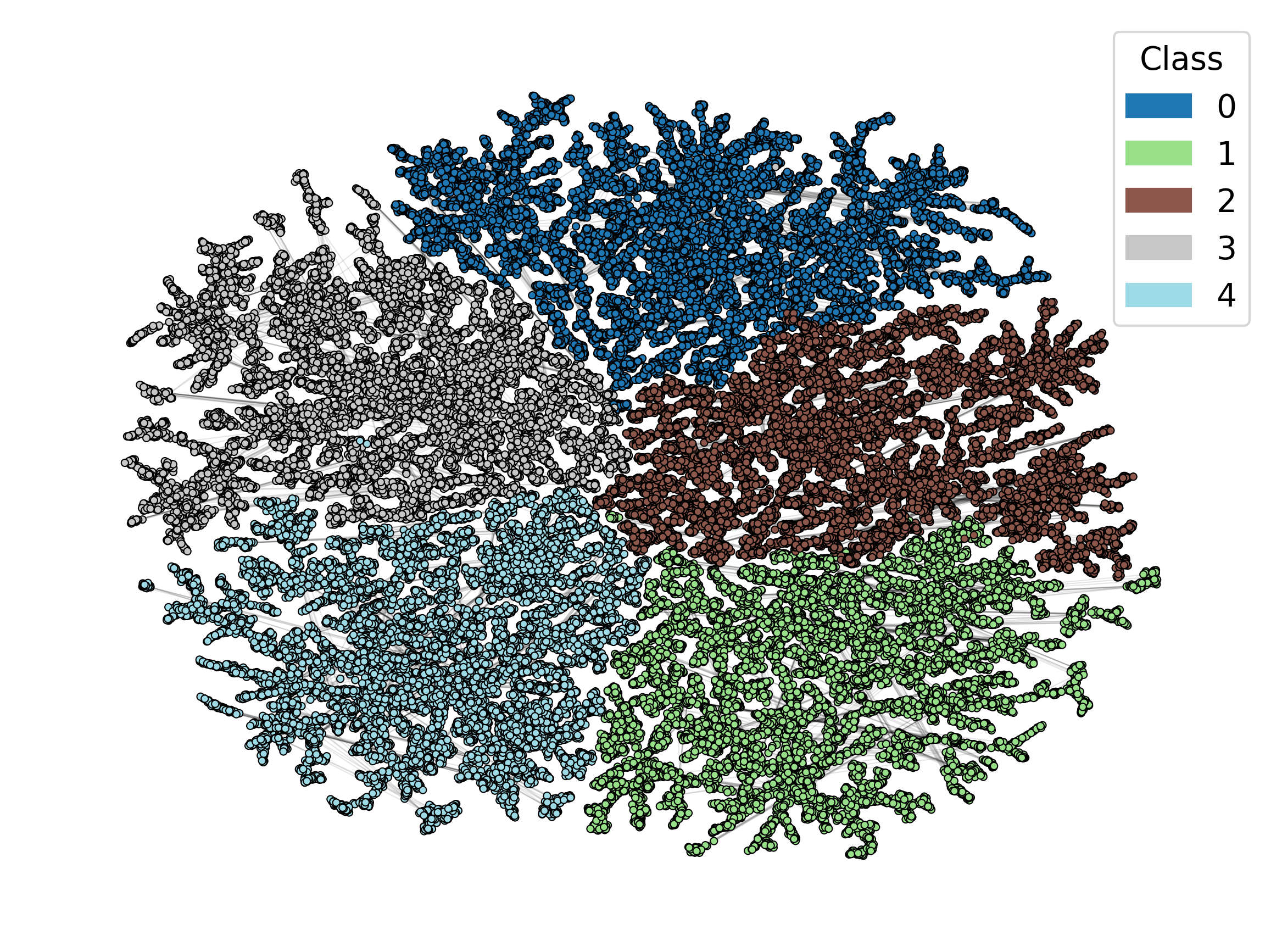}
    \caption{UMAP projection of an a-TMFG constructed for $N=100,000$ nodes generated from a Gaussian Markov Random Field (Section \ref{sec:gmrf}). The scalable algorithm successfully recovers the ground-truth clusters while preserving the dendritic, maximal planar topology characteristic of exact TMFGs.}
    \label{fig:largegraph}
\end{figure*}

As an illustration we show the TMFG estimated on a dataset of $N = 100000$ nodes with five clusters in Fig. \eqref{fig:largegraph}. The visualization layout was created by embedding the final TMFG with UMAP\cite{mcinnes_umap_2018}\footnote{NetworkX spring layout is comparatively significantly slower and unable to process large graphs.} as opposed to the original dataset $X$. The embedding of Gaussian factor model data typically result in blob like clustered graphs like in Fig \eqref{fig:factor_graph}. Similarly to Fig \eqref{fig:factor_tmfg} we observe a branching and dendritic like structure which captures the hierarchical dependencies embedded in the data. Clusters are recovered and their boundaries show minimal overlap illustrating the power of the method in building very large dependency graphs even when the input data was generated from a sparse GMRF model.

\subsection{$\alpha$'s impact on a-TMFG's performance} \label{sec:eval_alpha}

\begin{figure}[H]
    \centering
    \includegraphics[width=0.7\textwidth]{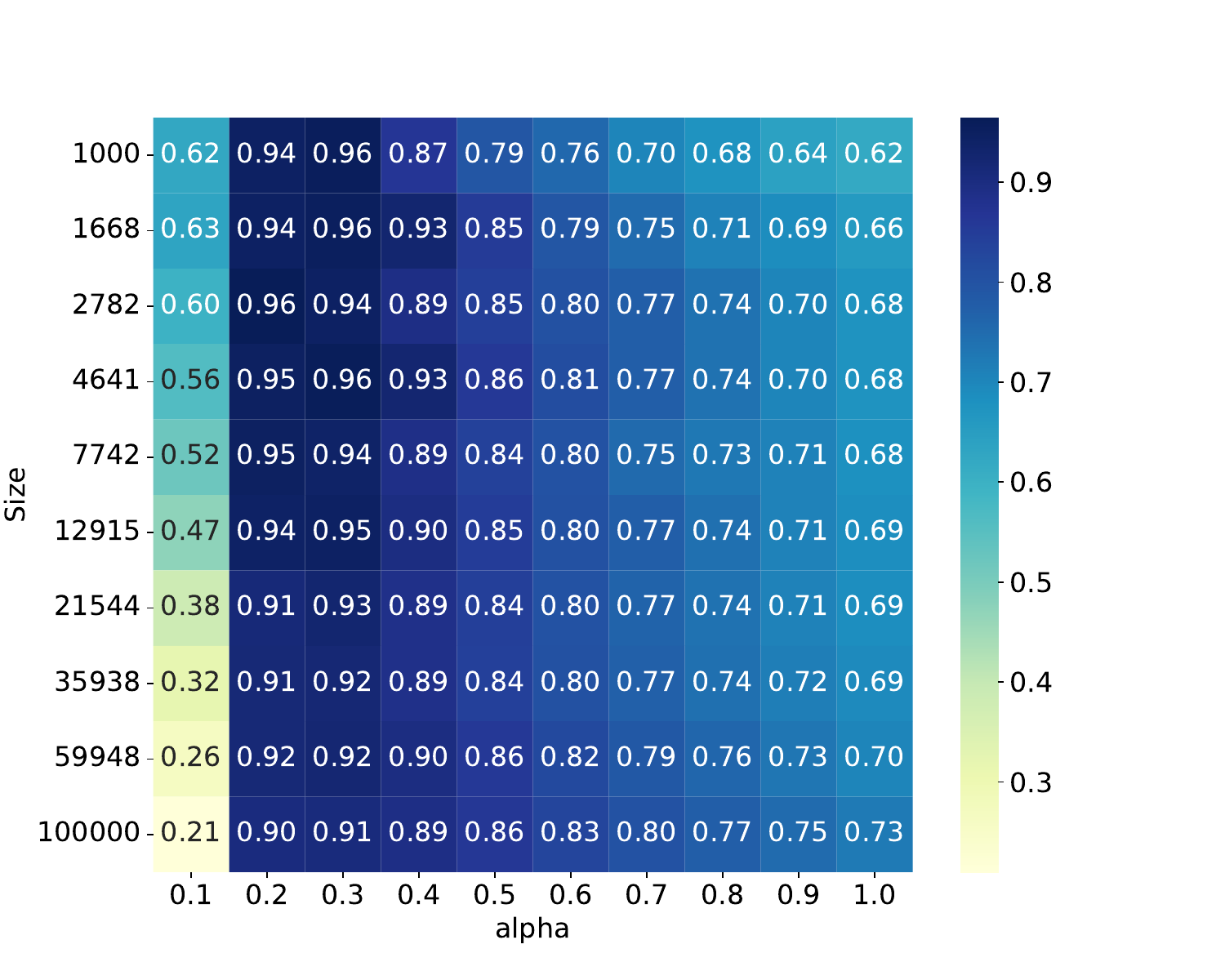}
    \caption{Average Jaccard similarity between a-TMFG and the exact connectivity matrix across varying dataset sizes $N$ and GMRF spatial dependence parameter $\alpha$. As analyzed in Section \ref{sec:eval_alpha}, the algorithm optimally recovers structures when $0.2 \leq \alpha \leq 0.3$.}
    \label{fig:heatmap}
\end{figure}

We generate synthetic datasets using the GMRF using the TMFG as the connectivity matrix of varying sizes ranging from $N=1000$ to $N=100000$ samples. We also vary $\alpha$ from 0.1 to 1.0 to investigate its impact on the dependencies embedded into the simulated data and by proxy its impact on the performance of our algorithm. Each size and alpha combination is sampled 5 times and we present the average Jaccard scores on a heat map in Fig. \eqref{fig:heatmap}.

We are interested in three regimes: 
\begin{enumerate}
    \item for $\alpha = 0.1$ the impact is significant and negative at all dataset sizes investigated but more pronounced at larger ones with Jaccard scores decreasing from 0.62 to 0.21 with increasing $N$.
    \item $\alpha \in (0.2, 0.3)$: Our algorithm is able to achieve Jaccard Scores above 0.90 at all sizes explored. This suggesting that suppressing long range dependencies by setting $\alpha < 2$ has a desirable impact on the embedded dependencies as seen by the performance of the TMFG as we predicted from Eq. \eqref{eq:gmrf_cov}.
    \item $\alpha \rightarrow 1$: The performance of the TMFG monotonously decreases as an increasing $\alpha$ directly impacts the term $\frac{\alpha^2}{4}$ in Eq. \eqref{eq:gmrf_cov} which eventually goes to 1 once $\alpha$ reaches 2. The effect is due to neighbors of neighbors dependencies becoming more prominent. The TMFG is designed to best capture 1-hop links between nodes within their direct neighborhood.
\end{enumerate}

\begin{figure}[H]
    \centering
    
    \begin{subfigure}{0.45\linewidth}
        \centering
        \includegraphics[width=\linewidth]{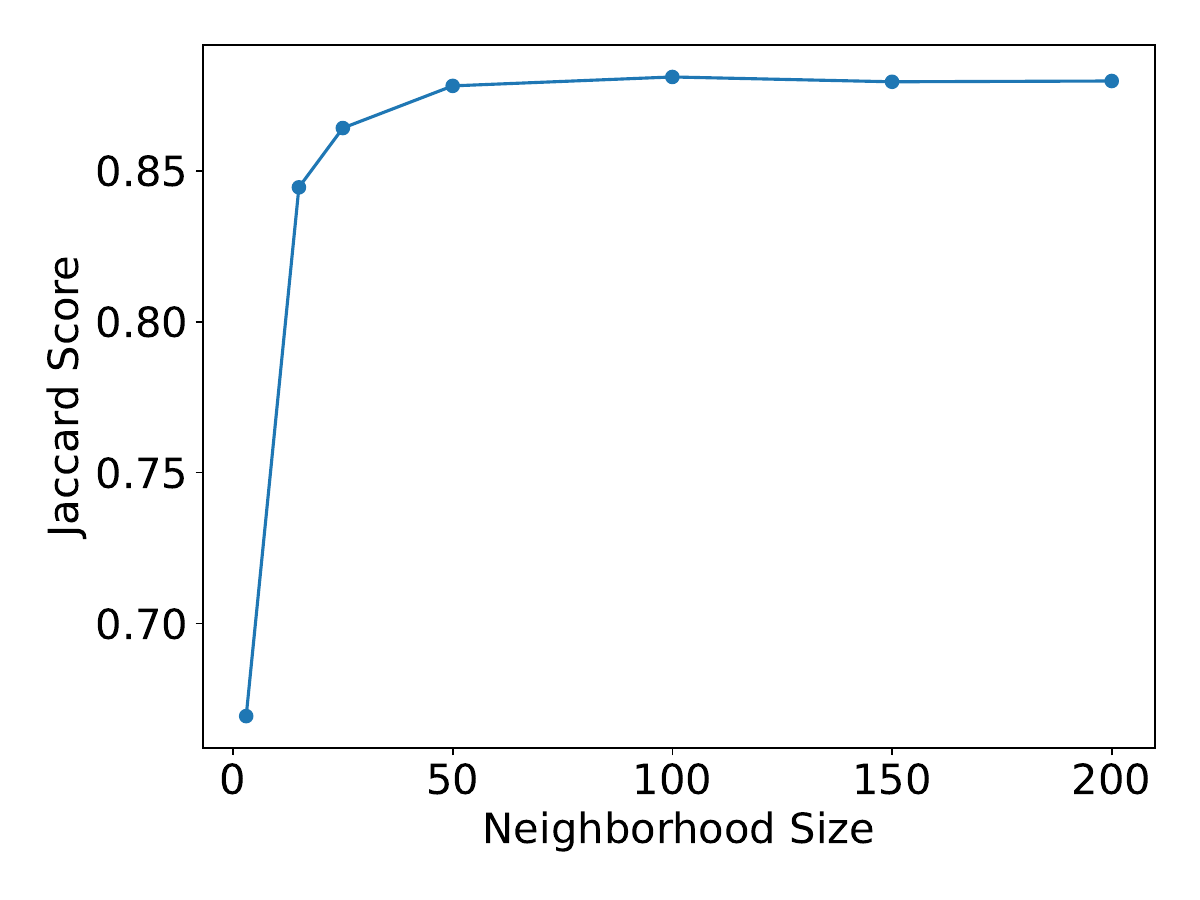}
        \caption{Jaccard Score Distribution}
        \label{fig:k_jaccard}
    \end{subfigure}
    ~
    \begin{subfigure}{0.45\linewidth}
        \centering
        \includegraphics[width=\linewidth]{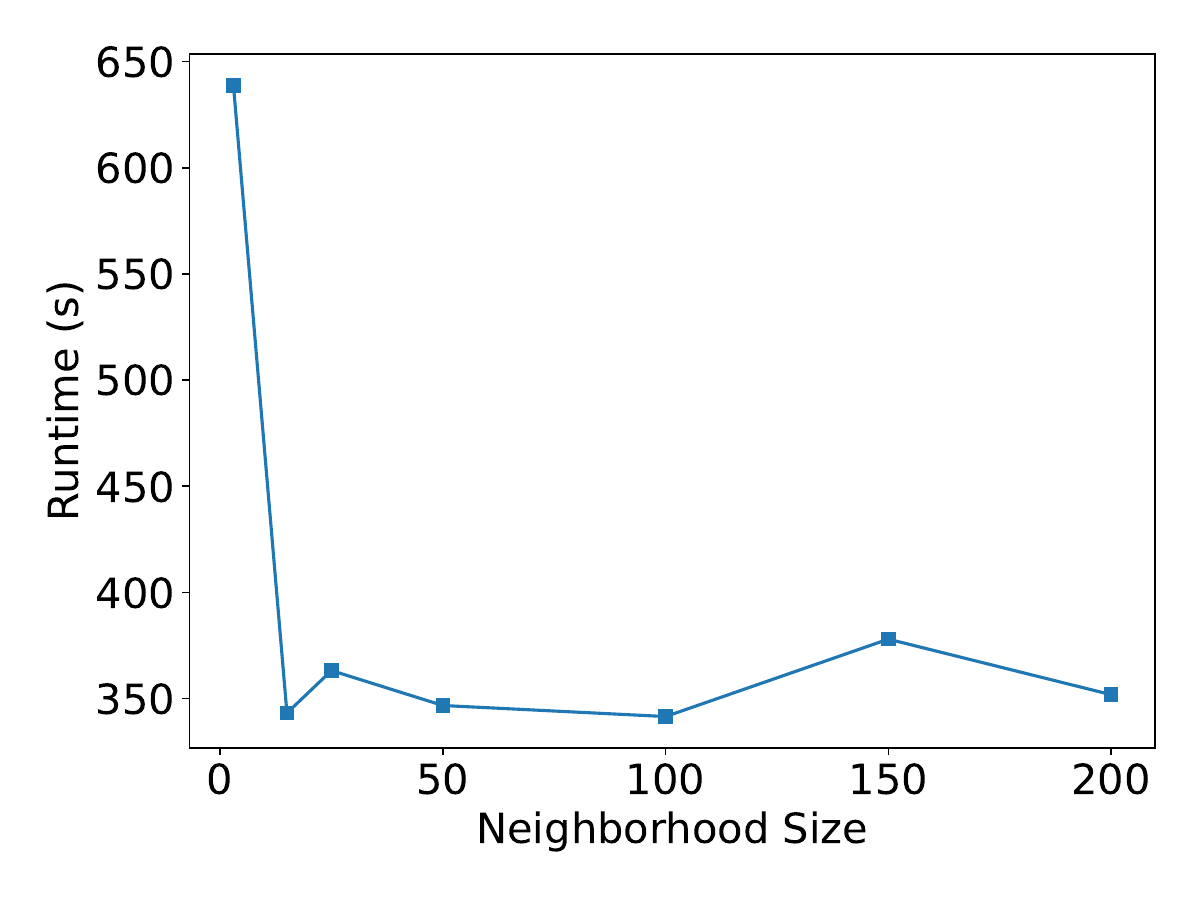}
        \caption{Runtime}
        \label{fig:k_runtime}
    \end{subfigure}
    
    \caption{Impact of the initial $k$-NN graph neighborhood size $k$ on a-TMFG construction for $N=50,000$. (\subref{fig:k_jaccard}) Jaccard similarity relative to the unconstrained graph. (\subref{fig:k_runtime}) Total runtime in seconds. A moderate neighborhood size ($k \ge 50$) is sufficient to achieve high structural fidelity while minimizing initial computational overhead (Section \ref{sec:eval_k}).}
    \label{fig:eval_k}
\end{figure}

\subsection{Neighborhood Size $k$} \label{sec:eval_k}

Next we show the impact of $k$ the neighborhood size of $G$ used as initialization knn graph in a-TMFG's routine. There is a starting cost to the computation of $G$ when $k$ is large that can be non-negligible which may be an incentive to prioritize smaller $k$ values. Additionally as datasets grow larger, utilizing smaller knn graphs allows us to control the scalability of graph construction techniques like a-TMFG. We generate datasets from the GMRF model\eqref{sec:gmrf} with $N=50000$ and $k=5$ and we construct TMFGs for a range of $k \in (3, 15, 25, 50, 100, 150, 200)$ values and we repeat the operation five times. An additional motivation for testing small $k$ values is due to the natural comparison of TMFG to knn graphs. The average degree of the TMFG is comparable to the average degree of a knn graph with $k=3$. Naturally we would like to understand the impact of initialization the algorithm with a small graph. In Fig. \eqref{fig:k_jaccard} we show the average Jaccard Score as we increase $k$. In Fig. \eqref{fig:k_runtime} we show the impact of increasing $k$ on the construction of the TMFG.

\subsection{Face Universe Size $\mathcal{F}$} \label{sec:eval_universe}

% We investigate the impact of varying $|\mathcal{F}| \in [500, 1000, 5000, 7500, 10000, 25000, 50000, 75000, 100000] $. Our motivation is to explore the impact of very small face universe sizes to large keeping in mind that

% \subsection{Face Universe Size $\mathcal{F}$} \label{sec:eval_universe}

We investigate the impact of varying the maximum face universe limit, $|\mathcal{F}| \in [500,...,100000]$. Our motivation is to explore the trade-off between strict memory constraints, runtime efficiency, and the structural fidelity of the resulting graph. 

\begin{figure}[H]
    \centering
    
    \begin{subfigure}{0.45\linewidth}
        \centering
        \includegraphics[width=\linewidth]{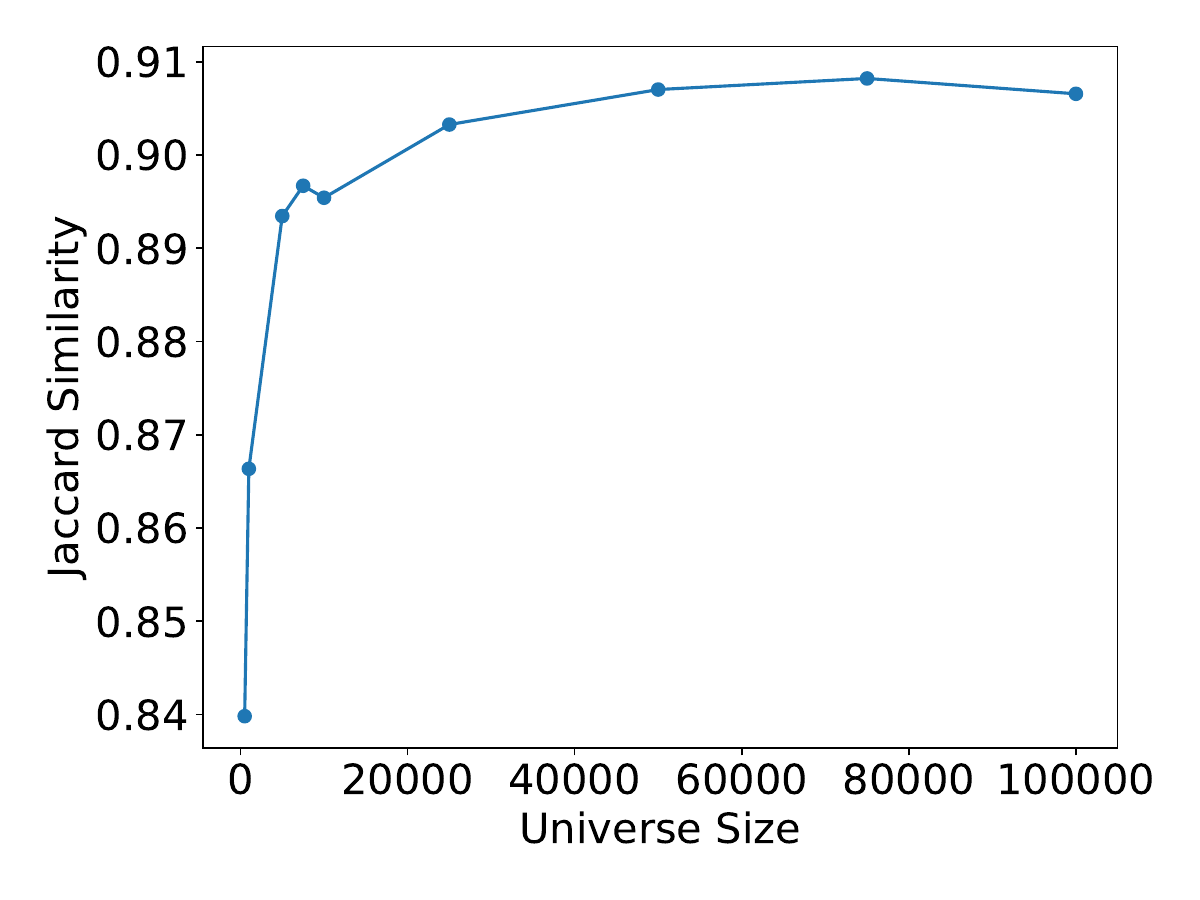}
        \caption{Jaccard Score Distribution}
        \label{fig:f_jaccard}
    \end{subfigure}
    ~
    \begin{subfigure}{0.45\linewidth}
        \centering
        \includegraphics[width=\linewidth]{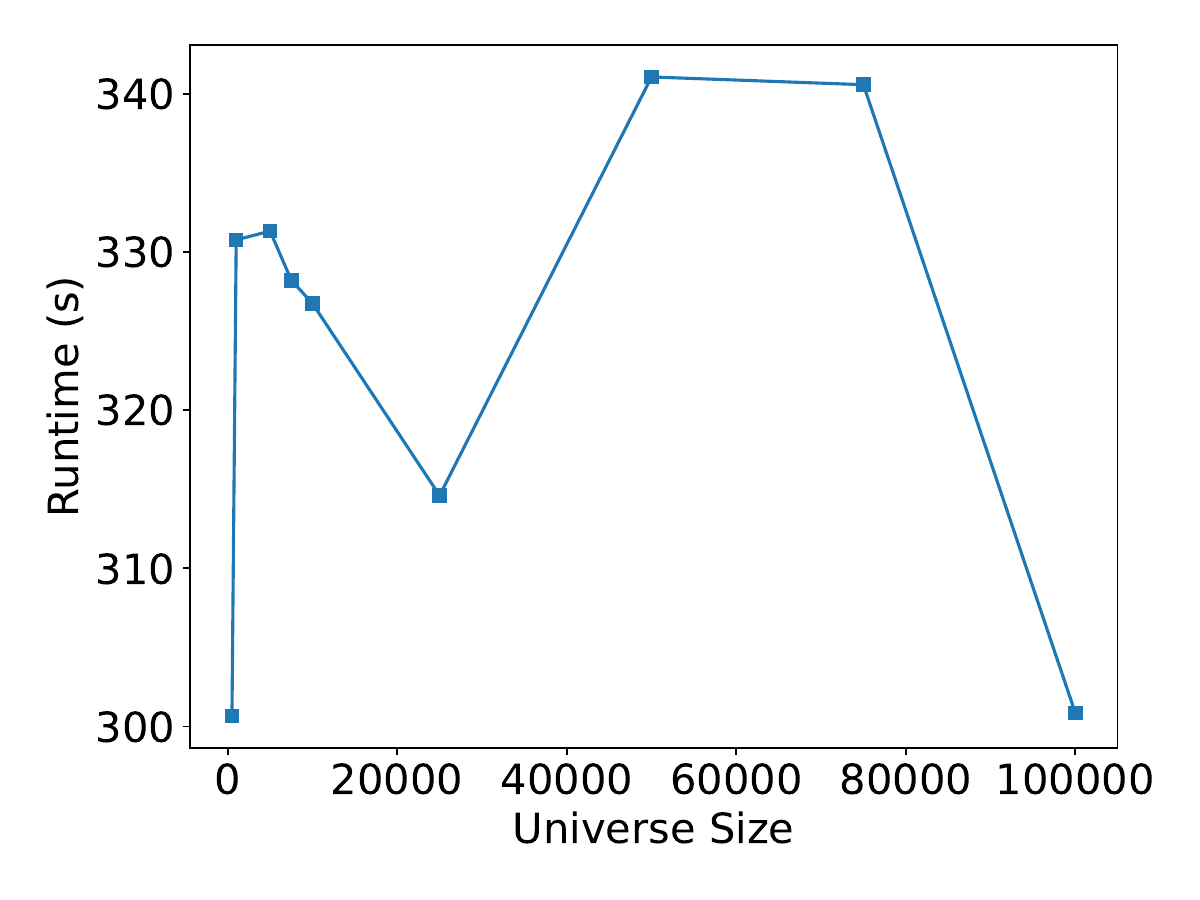}
        \caption{Runtime}
        \label{fig:f_runtime}
    \end{subfigure}
    
    \caption{Trade-offs associated with the maximum face universe limit $|\mathcal{F}|$ for $N=100,000$. (\subref{fig:f_jaccard}) Jaccard similarity to the exact TMFG. (\subref{fig:f_runtime}) Construction runtime. Constraining $|\mathcal{F}|$ to the ``elbow" of the curve in the Jaccard similarity score of Figure (\subref{fig:f_jaccard}) optimally balances structural accuracy with computational and memory efficiency (Section \ref{sec:eval_universe}).}
    \label{fig:universe}
\end{figure}

In Fig. \ref{fig:f_jaccard}, we observe the impact of the universe limit on the Jaccard similarity between the a-TMFG and the unconstrained exact graph. When $|\mathcal{F}|$ is severely restricted (e.g., under 5,000), the algorithm is forced to aggressively prune active faces. When the local search naturally exhausts, the global rescue mechanism can only query the HNSW index using this severely truncated "memory" of the graph's frontier. This limitation leads to suboptimal topological jumps and subsequently lower Jaccard scores. 

However, as the universe size expands, the Jaccard score rises steeply, ultimately plateauing near 0.91 once $|\mathcal{F}|$ reaches approximately 25,000. This plateau empirically validates our core hypothesis from Section \ref{sec:scalability}: the algorithm does not need to store the entire $\mathcal{O}(N^2)$ history of the graph's construction. Retaining only a sliding window of the most recent active exploration frontier is sufficient to approximate the exact TMFG with exceptionally high fidelity.

In Fig. \ref{fig:f_runtime}, the runtime exhibits a non-linear, bounded response to the universe size. At extremely low limits, the engine incurs continuous computational overhead from triggering the garbage collection (pruning) routine and executing frequent HNSW global rescues. Conversely, at very high limits, the priority queue becomes massively bloated—increasing the logarithmic cost of standard heap operations—and batch HNSW queries scale into heavy matrix operations. 

These results offer a clear heuristic for practitioners: the optimal universe limit lies at the "elbow" of the fidelity curve (in this experiment, roughly $0.2N$ to $0.5N$). Setting $|\mathcal{F}|$ within this regime maximizes structural accuracy while maintaining a lean memory footprint and circumventing the computational bloat of an unbounded face universe.

\subsection{Runtime Scaling and Complexity} \label{sec:eval_runtime}

\begin{figure}[H]
    \centering
    \includegraphics[width=0.75\textwidth]{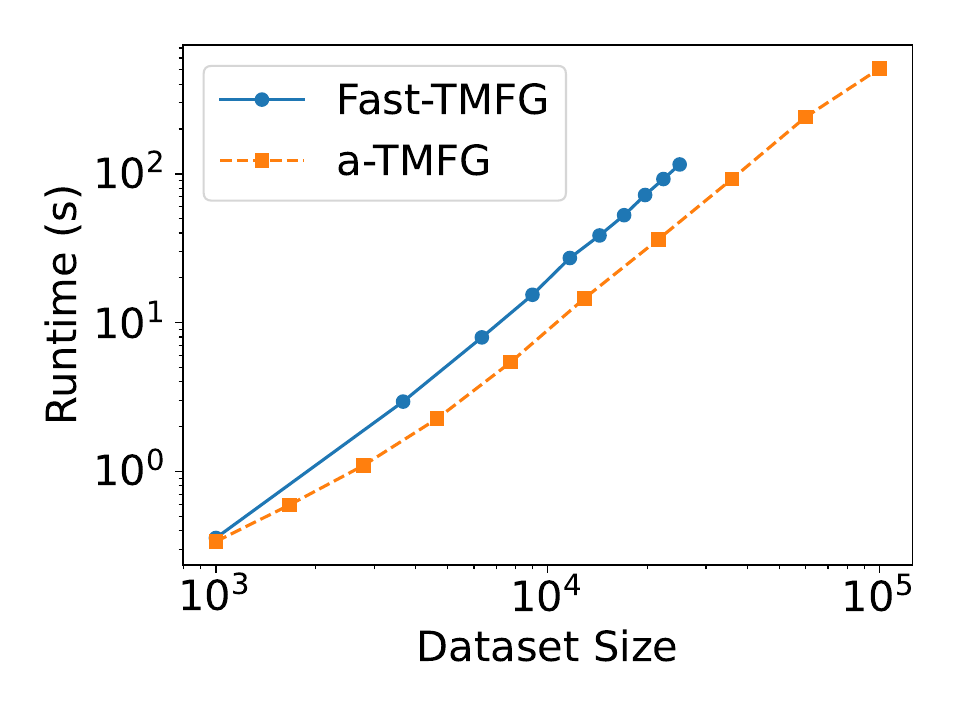}
    \caption{Empirical runtime scaling of the proposed a-TMFG against the exact Fast-TMFG algorithm. The bounded universe and HNSW queries enable a-TMFG to maintain a near-linear $\mathcal{O}(UN)$ trajectory, whereas exact methods hit computational bottlenecks near $N=25,000$ due to their inherent $\mathcal{O}(N^2)$ complexity (Section \ref{sec:eval_runtime}).}
    \label{fig:runtime}
\end{figure}

To empirically validate the theoretical efficiency gains of our approach, we compare the execution time of a-TMFG against the exact Fast-TMFG algorithm\cite{briola2022dependency}. Figure \ref{fig:runtime} illustrates the runtime, measured in seconds, as a function of the dataset size $N$.

The results demonstrate a stark contrast in scalability. The Fast-TMFG algorithm exhibits a steep, non-linear growth profile characteristic of its underlying $\mathcal{O}(N^2)$ computational and memory constraints. As observed in the benchmark, this quadratic scaling renders the traditional approach practically intractable for datasets extending beyond $N \approx 25,000$ on standard hardware. 

In contrast, a-TMFG leverages the bounded face universe, lazy-deletion priority queue, and efficient HNSW index queries to maintain a significantly flatter, near-linear growth trajectory. This controlled complexity ($\approx \mathcal{O}(UN)$) enables our method to effortlessly construct a complete maximal planar graph for $N=100,000$ nodes in just over 500 seconds—a fraction of the time it would theoretically take exact methods to process a quarter of that volume. Consequently, a-TMFG unlocks the ability to apply topological filtering to massive tabular datasets with highly efficient memory use, bypassing the need for heavy distributed computing resources.

\section{Discussion and Conclusion} \label{sec:discussion}

In this paper, we introduced the Approximate Triangular Maximally Filtered Graph (a-TMFG), a highly scalable framework designed to estimate complex topological dependency graphs from data. Traditional TMFG construction has historically been bottlenecked by $\mathcal{O}(N^2)$ memory and runtime complexities due to its reliance on dense correlation matrices. By integrating approximate nearest neighbor indexing (HNSW), a bounded face-universe, and lazy-deletion priority queues, a-TMFG effectively reduces this complexity to approximately $\mathcal{O}(UN)$, where $U$ is a strictly controlled universe limit.Our evaluations on synthetic Gaussian Markov Random Field (GMRF) datasets demonstrate that a-TMFG successfully recovers ground-truth hierarchical structures and intrinsic cluster boundaries. Notably, as demonstrated in our large-scale manifold embedding ($N=50,000$), the algorithm preserves the mathematically desirable planar and dendritic properties of the TMFG without collapsing into dense, uninformative topologies. 

Furthermore, runtime comparisons establish that a-TMFG comfortably processes large-scale datasets that force exact and parallelized TMFG methods into computational intractability.Despite these advances, the current methodology presents certain limitations. The reliance on a localized $k$-NN initialization means that the algorithm is sensitive to the choice of the neighborhood parameter $k$. If $k$ is set too low, the algorithm relies heavily on the global rescue mechanism, which, while effective, can alter the local dependency structures if the dataset is highly fragmented. Additionally, because a-TMFG is a greedy approximation, it trades a marginal degree of exact structural fidelity for its exponential speedup.

Future work will focus on three primary avenues. First, adaptive heuristics for dynamically tuning the universe limit $\mathcal{F}$ and neighborhood size $k$ during runtime could further optimize the balance between speed and accuracy. Second, applying a-TMFG to real-world tabular datasets across domains such as finance, physics, biology and chemistry will be essential to validate its utility in empirical settings. Finally, investigating the downstream performance of a-TMFG adjacency matrices as inputs for Graph Neural Networks (GNNs) remains a highly promising direction, potentially bridging the gap between tabular data and modern graph representation learning.

\section{Acknowledgments} \label{sec:ack}

I thank Tim Gebbie for helpful discussions, proof reading and editing.  

\newpage

\bibliographystyle{apalike-ejor} %plainurl unsrtnat IEEEtrans apalike
\bibliography{bibzotero}

\newpage
\appendix

\section{Evaluation Metrics}

\subsection{Weighted Average Shortest Path Within Partitions.}

Given a graph $G = (V, E)$ and a partition of its nodes into groups, we define a metric that measures the internal connectivity of each group by computing the average shortest path within each group and then performing a weighted sum.
For each group $C_k$ in the partition $\mathcal{C}$, let $d_{ij}$ be the shortest path distance between nodes $i$ and $j$ within the group. The average shortest path for group $C_k$ is:
\begin{equation}
    L(C_k) = \frac{1}{|C_k|(|C_k|-1)} \sum_{i,j \in C_k, i \neq j} d_{ij}.
\end{equation}

To obtain an overall metric, we compute the weighted sum of these values, with weights as a function of the group sizes:
\begin{equation}
    L_{\text{weighted}} = \sum_{k} w_k L(C_k),
\end{equation}
where the weight $w_k$ is given by:
\begin{equation}
    w_k = \frac{|C_k|}{\sum_{m} |C_m|}.
\end{equation}

\end{document}